\documentclass[journal,twoside,web]{ieeecolor}
\usepackage{tmi}
\usepackage{cite}
\usepackage{amsmath,amssymb,amsfonts}
\usepackage{algorithmic}
\usepackage{graphicx}
\usepackage{textcomp}
\usepackage{tipa}
\usepackage{bbding}
\usepackage{graphicx}
\usepackage{multirow}
\usepackage{color, colortbl}
\usepackage{float}
\usepackage{amsmath}
\usepackage{amssymb}
\usepackage{booktabs}
\usepackage{colortbl}
\usepackage{xcolor}

\newcommand{\eg}{e.g.}
\newcommand{\ie}{i.e.}
\newcommand{\etal}{et al.}

\def\BibTeX{{\rm B\kern-.05em{\sc i\kern-.025em b}\kern-.08em
    T\kern-.1667em\lower.7ex\hbox{E}\kern-.125emX}}
\begin{document}
\title{Cross-Modal Conditioned Reconstruction for Language-guided Medical Image Segmentation}
\author{Xiaoshuang Huang, Hongxiang Li, Meng Cao, Long Chen, \IEEEmembership{Member, IEEE}, Chenyu You, and Dong An
\thanks{This work was supported in part by the National Key Research and Development Program of China (2022YFF0608404). (Corresponding author: Dong An) }
\thanks{ Xiaoshuang Huang, Dong An are with the
College of Information and Electrical Engineering, China Agricultural University, Beijing, 100083, China and also with the Beijing Engineering and Technology Research Center for Internet of Things in Agriculture, China Agricultural University, Beijing 100083, China (e-mail: huangxiaoshuang@cau.edu.cn; andong@cau.edu.cn).}
\thanks{ Hongxiang Li is with the
School of Electronic and Computer Engineering, Peking University, Shenzhen, 518055, China (e-mail: lihongxiang@stu.pku.edu.cn).}
\thanks{ Meng Cao is with the
Mohamed bin Zayed University of Artificial Intelligence, Masdar City, Abu Dhabi, United Arab Emirates (e-mail: mengcaopku@gmail.com).}
\thanks{ Long Chen is with the
Department of Computer Science and Engineering
School of Engineering, The Hong Kong University of Science and Technology, Hong Kong, 999077, China (e-mail: longchen@ust.hk).}
\thanks{ Chenyu You is with the Department of Electrical Engineering, Yale University, New Haven, CT, USA. (e-mail: chenyu.you@yale.edu).}
}

\maketitle

\begin{abstract}
Recent developments underscore the potential of textual information in enhancing learning models for a deeper understanding of medical visual semantics. However, language-guided medical image segmentation still faces a challenging issue. Previous works employ implicit and ambiguous architectures to embed textual information. This leads to segmentation results that are inconsistent with the semantics represented by the language, sometimes even diverging significantly. To this end, we propose a novel cross-modal conditioned Reconstruction for Language-guided Medical Image Segmentation (RecLMIS) to explicitly capture cross-modal interactions, which assumes that well-aligned medical visual features and medical notes can effectively reconstruct each other. We introduce conditioned interaction to adaptively predict patches and words of interest. Subsequently, they are utilized as conditioning factors for mutual reconstruction to align with regions described in the medical notes.
Extensive experiments demonstrate the superiority of our RecLMIS, surpassing LViT by 3.74\% mIoU on the publicly available MosMedData+ dataset and achieving an average increase of 1.89\% mIoU for cross-domain tests on our QATA-CoV19 dataset. Simultaneously, we achieve a relative reduction of 20.2\% in parameter count and a 55.5\% decrease in computational load. The code will be available at https://github.com/ShawnHuang497/RecLMIS.
\end{abstract}


\section{Introduction}
\label{sec:intro}

Medical image segmentation (MIS) is one of the most fundamental yet challenging tasks in medical image analysis. 
\begin{figure}[H]
    \centering
    \includegraphics[width=\linewidth]{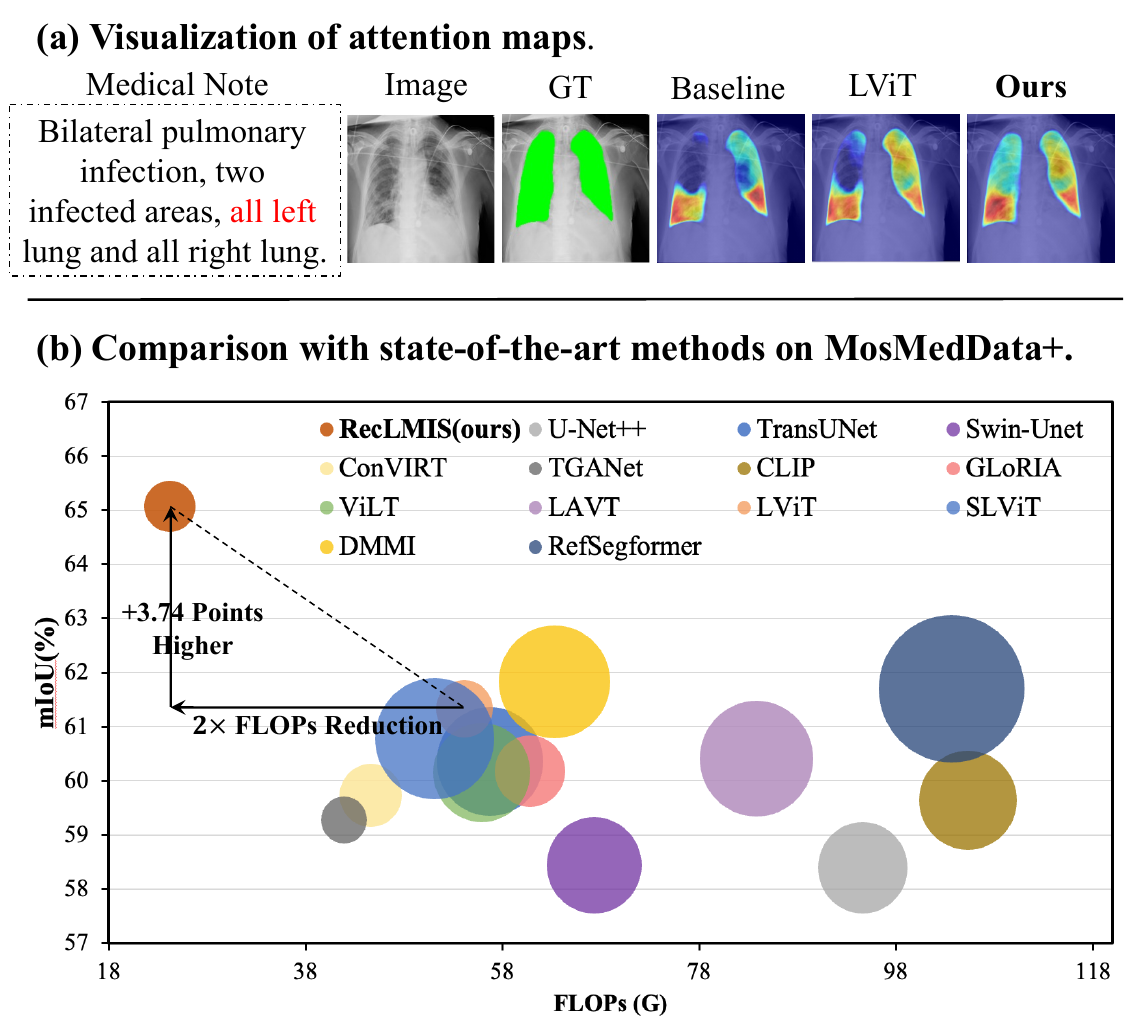}
    \caption{\textbf{(a)} Existing methods (\eg, LViT\cite{li2023lvit}) face the issue of not fully and effectively adhering to the correct text prompts. For example, the text 'all left' is not fully reflected in the attention maps in LViT~\cite{li2023lvit}. Compared to that, the RecLMIS we proposed can focus on regions that match the text prompts properly. 
    \textbf{(b)} Comparison with state-of-the-art methods on MosMedData+~\cite{morozov2020mosmeddata, hofmanninger2020automatic, li2023lvit} dataset on mIoU (y-axis), parameter count (size of the area), and FLOPs (x-axis). The RecLMIS we proposed is superior in both performance and FLOPs.}
    \label{fig: motivation}
\end{figure}
\begin{figure*}[ht]
\centerline{\includegraphics[width=\linewidth]{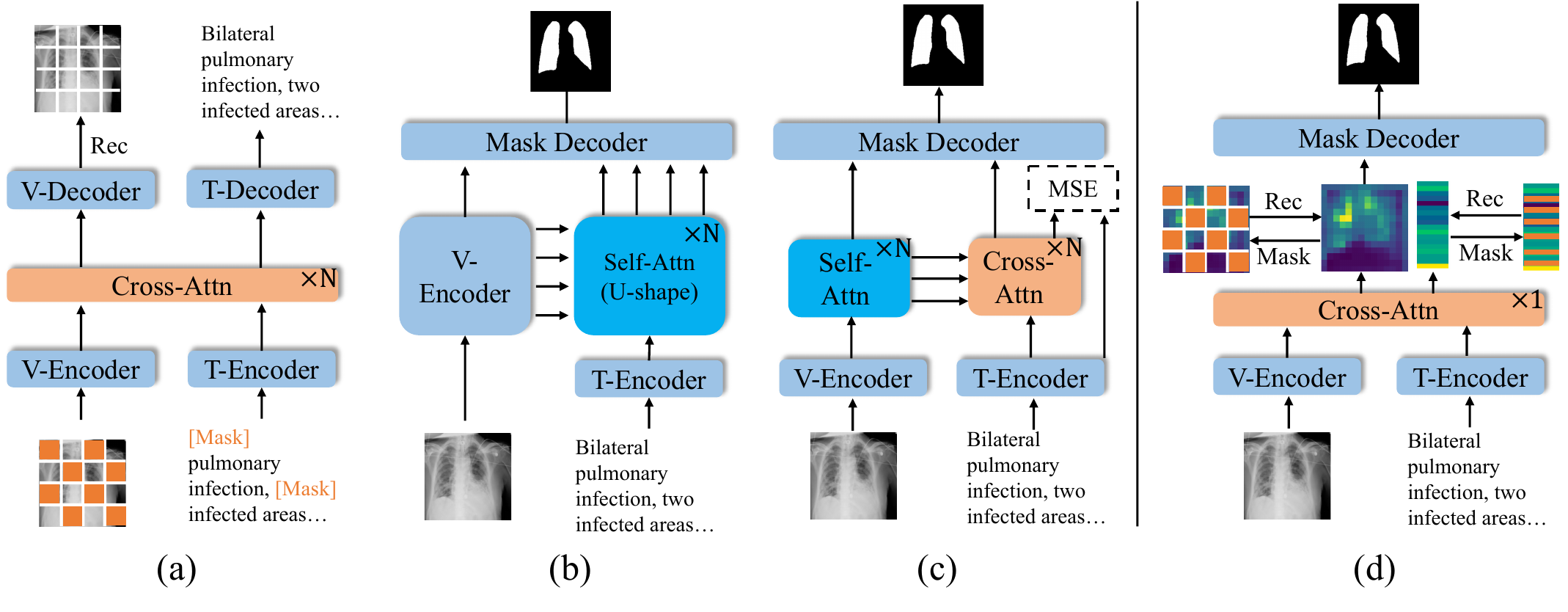}}
\caption{A comparative analysis of different model architectures for mask modeling and Language-guided Medical Image Segmentation (LMIS). The orange patch represents the masked region. The 'V-Encoder', 'T-Encoder', 'V-Decoder', and 'T-Decoder', represent vision encoder, text encoder, vision decoder, and text decoder, respectively. 'Attn' and 'MSE' stand for 'Attention' and Mean Squared Error, respectively. 
\textbf{(a)} The MAE and BERT series mask modeling architectures involve masking the original image/sentence and subsequently using the vision decoder V-Decoder/T-Decoder for reconstruction. For achieving referring image segmentation, fine-tuning downstream task datasets is necessary. This approach encompasses methods such as~\cite{chen2022multi,wang2023swinmm,devlin2018bert}.
\textbf{(b)} The parallel U-shape architecture (LViT~\cite{li2023lvit}, LAVT~\cite{yang2022lavt} integrates an additional parallel U-shape structure to the original segmentation model for processing and fusing text, aiming to align visual and textual features.
\textbf{(c)} The dual-branch fusion architecture~\cite{liu2023multi,hu2023beyond} employs N layers of cross-attention in the text branch to align visual and textual features, minimizing the Mean Squared Error (MSE) between post-interaction and pre-interaction textual features.
\textbf{(d)} Our cross-modal conditional reconstruction fusion architecture proposes using textual/visual features and conditions to reconstruct current visual/textual features during training, enabling the alignment of visual and textual features with just a single layer of cross-attention.} \label{fig:arch_compare}
\end{figure*}
It plays a crucial role in numerous medical applications, such as disease identification~\cite{liu2019comparison}, diagnosis~\cite{aggarwal2021diagnostic}, prognosis~\cite{tran2021deep}, and so on. However, owing to the typically expensive and challenging nature of obtaining expert-level segmentation annotations for medical images, there remains a scarcity of high-quality annotations in medical image segmentation. A feasible solution is to utilize medical notes to remedy the quality defects of medical image segmentation datasets~\cite{li2023lvit}.
In this paper, we follow~\cite{li2023lvit} and focus on language-guided medical image segmentation, which is a challenging vision-language task.

Recently, visual language representation learning has achieved great success, \eg, CLIP~\cite{radford2021learning}, ALIGN~\cite{jia2021scaling}, and Florence~\cite{yuan2021florence}. The core idea is to project images and text into a common potential space according to the semantic similarities of image-text pairs. By moving from single visual modality to multi-modality, significant progress has advanced in many computer vision tasks such as image segmentation~\cite{luddecke2022image, hu2023beyond}, image generation~\cite{crowson2022vqgan, zhang2023adding}, and the video grounding~\cite{luo2022clip4clip, ko2023meltr}. However, how to leverage language to guide medical image segmentation is not well explored. 

The one challenge is how to capture discriminative multi-modal representations for medical image segmentation. Although several recent approaches~\cite{li2023lvit, tomar2022tganet} have attempted to incorporate textual features to improve medical image segmentation, they all rely on ambiguous attention mechanisms to embed text into visual features. Due to the highly similar foreground and background in medical images, this implicit modeling often leads to segmentation results that do not match the descriptions in medical notes. As shown in Fig.~\ref{fig: motivation}(a), we present the attention maps of the visual encoder. It is evident that the previous state-of-the-art method, LViT~\cite{li2023lvit}, struggled to accurately attend to the semantic content of ``all left'' in the text, leading to incorrect prediction. 

Another challenge lies in the relatively slow inference speed. Some methods achieve better integration of textual and visual semantics by introducing specific neural network layers. While this enhances performance, it also implies a greater number of parameters and computational load, consequently leading to slower inference speeds for the model. 
For instance, a dual-layer parallel U-shape structure in Fig.~\ref{fig:arch_compare}(b) is introduced to integrate textual semantics in LViT~\cite{li2023lvit}. Moreover, some Referring Image Segmentation (RIS) methods~\cite{liu2023multi,hu2023beyond} in the natural world, as shown in Fig.~\ref{fig:arch_compare}(c), also introduce multi-layer structures and complex interactions to fuse visual and linguistic features, which increase inference time and parameter count. This will be elaborated on in detail in Sec.~\ref{sec: efficiency}).

To address this issue, we propose a novel cross-modal conditioned \textbf{Rec}onstruction for \textbf{L}anguage-guided \textbf{M}edical \textbf{I}mage \textbf{S}egmentation (\textbf{RecLMIS}) as shown in Fig.~\ref{fig:arch_compare}(d), which introduces conditioned reconstruction to explicitly capture cross-modal interactions of medical image and text.
Concretely, we assume that aligned text words and image regions reconstruct each other well.
We first utilize the Conditional Interaction Module to score each token to adaptively capture patches of interest ($W_{poi}$) and words of interest ($W_{woi}$).
Then, we treat language-guided medical image segmentation as a cross-modal reconstruction to explicitly investigate semantically relevant visual and textual concepts.
$W_{poi}$/$W_{woi}$ are taken as probability distributions to mask images/texts randomly, and we consider $W_{woi}$/$W_{poi}$ as soft conditions to reconstruct image/text features.
Finally, we use the mean squared error (MSE) to measure the similarity between the reconstructed image/text and the original image/text features. 
Simultaneously, we introduce a novel conditional contrastive learning loss to leverage $W_{poi}$ and $W_{woi}$ to dominate cross-modal contrasts.

Notably, compared to previous works~(\eg, LViT~\cite{li2023lvit}), RecLMIS benefits from the explicit alignment of cross-modal features, so that complex multi-level interactive alignment is not required, and the proposed reconstruction module is removable during the inference, thus reducing the computational load and parameter count as shown in Fig.~\ref{fig: motivation}(c).
Extensive experiments on widely used benchmark datasets prove the superiority of our RecLMIS.
Our contributions can be summarized as follows:
\begin{itemize}
    \item We have proposed the conditioned interaction module. It focuses on the key feature token information aligned between images and text, and we have demonstrated that can reduce interference from background information present in both images and text.
    \item We have introduced the Conditioned Language Reconstruction module and Conditioned Vision Reconstruction. Both of them utilize cross-modal masking features for reconstruction, and we have demonstrated that could enhance the representational capacity of the visual encoder.
    \item We propose an efficient medical segmentation method, \textbf{RecLMIS}, which employs conditioned contrastive learning and cross-modal conditioned reconstruction to learn fine-grained cross-modal alignment. Extensive experiments show that the proposed RecLMIS outperforms the state-of-the-art methods on widely used public benchmarks, exhibiting more accurate results and faster inference speed. 
\end{itemize}

\section{Related Work}
\label{sec:related_work}
\subsection{Medical Image Segmentation.}
With the advent of deep convolutional neural networks (CNN), the U-shape architecture was introduced for medical image segmentation in~\cite{ronneberger2015u}, which is improved by U-Net++~\cite{zhou2018unet++}, UNet3+~\cite{huang2020unet}, nnUnet~\cite{isensee2021nnu}, and so on. 
Recently, driven by the success of the Transformer and vision transformer, the researchers introduced two advanced combining CNN with transformer and pure transformer models. For example, Chen \etal~\cite{chen2021transunet} exploit both detailed high-resolution spatial semantic information from CNN features and the global context encoded by Transformers. Cao \etal~\cite{cao2022swin} introduce a pure Transformer-based U-shape Encoder-Decoder architecture with a skip connection. Furthermore, there is ongoing research~\cite{yao2022transclaw, zhang2021transfuse, chen2023transattunet, yuan2023effective, huang2022missformer} that delves into the development of specialized medical image Transformer structures, considering both encoder and decoder perspectives and attain notable performance. One of the pivotal factors contributing to the success of these models is that Transformer has an expansive global receptive field, endowing them with heightened flexibility and enhanced effect in addressing intricate patterns and relationships. 

The method most related to ours is LViT~\cite{li2023lvit}, which proposed a text-augmented medical image segmentation model based on UNet. Differently from~\cite{li2023lvit}, we only fuse text and visual features in the last visual encoder stage, with text and vision conditional reconstruction for comprehensive cross-modal understanding of the text expression and vision context.

\begin{figure*}[ht]
\centerline{\includegraphics[width=1.0\textwidth]{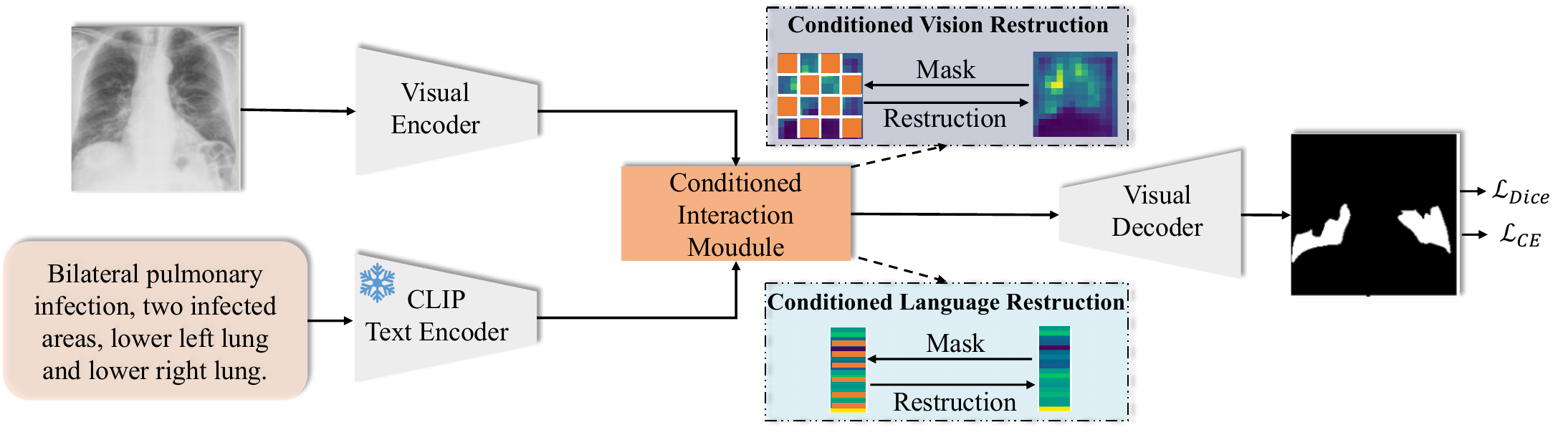}}
\caption{Overview of the proposed cross-modal conditioned \textbf{Rec}onstruction for \textbf{L}anguage-guided \textbf{M}edical \textbf{I}mage \textbf{S}egmentation (\textbf{RecLMIS}). Given a pair of medical images and notes as prompts, we first exploit the visual encoder and text encoder to extract image and text features respectively. The Conditioned Interaction (Sec.~\ref{sec: ci}) module is designed to align features from both vision and language inputs. The conditioned vision-language reconstruction encompasses a Conditioned Vision Reconstruction module (CVR, Sec.~\ref{sec:CVR}) and a Conditioned Language Reconstruction module (CLR, Sec.~\ref{sec:CLR}), serving the purposes of medical image reconstruction and language feature reconstruction, respectively. Finally, a lightweight mask predictor is employed to segment the designated region indicated by the text prompts.
} \label{model}
\end{figure*}

 \subsection{Referring Image Segmentation} 
Compared to traditional single-modality visual segmentation tasks that operate under fixed category conditions, Referring Image Segmentation (RIS) must deal with the significantly more extensive lexicons and syntactic variations inherent in human natural languages. In conventional pipelines, there exist two primary processes: (1) extracting text features with language Transformers or recurrent neural networks, and extracting vision features with vanilla fully convolutional networks, and (2) building mask predictors with the fusion of these multi-modal features. 

The second process is the central component that previous research has emphasized. Liu \etal~\cite{liu2017recurrent} propose a multi-modal LSTM, which models each word in every recurrent stage to fuse the word feature with vision features. Shi \etal~\cite{shi2018key}, Chen \etal~\cite{chen2019see}, Ye \etal~\cite{ye2019cross}, and Hu \etal~\cite{feng2021bidirectional} employ different attention mechanisms to model cross-modal relationships between language and visual features. Huang et al.~\cite{huang2020referring} leverage knowledge about sentence structures to capture different concepts (\eg, categories, attributes, relations, etc.) in multi-modal features, whereas Hui \etal~\cite{hui2020linguistic} introduces text structure-guided context modeling to facilitate the analysis of text structures for enhanced language comprehension. Ding \etal~\cite{ding2021vision} design a Transformer decoder for fusing text and visual features.

In addition to that, many researchers have gradually focused on the first process, which has also achieved better performance, such as LAVT~\cite{yang2022lavt, wu2024towards} and DMMI~\cite{hu2023beyond}. The former designs an early fusion scheme that effectively harnesses the Transformer encoder for modeling multi-modal context, and the latter reconstructs the masked entity phrase conditioned on the visual feature for facilitating the comprehensive understanding of the text expression. Moreover, many recent works~\cite{xie2024satr,liu2023multi,shah2024lqmformer,chng2024mask} have proposed complex feature fusion modules to achieve significant performance improvements. For example, Liu et al.~\cite{liu2023multi} propose a multi-modal mutual attention module and Shah et al.~\cite{shah2024lqmformer} develop LQMFormer with the gaussian-enhanced multi-modal fusion module.
While visual-language segmentation models have made remarkable advancements in the realm of natural images, their direct application to medical image segmentation proves unsatisfactory due to the significant divergence between medical and natural images. 

\section{Method}
\label{sec:method}
Our RecLMIS adopts the encoder-decoder paradigm as shown in Fig.~\ref{model}. 
The fundamental architecture for both encoder and decoder is based on the U-Net~\cite{ronneberger2015u} structure with the PLAM~\cite{li2023lvit} module. 
After the feature encoder, the Conditioned Interaction (Sec.~\ref{sec: ci})  module with conditioned contrastive learning loss is employed for aligning visual and text features. Meanwhile, the Conditioned Language Reconstruction module (Sec.~\ref{sec:CLR}) with vision-to-text reconstruction loss, and the Conditioned Vision Reconstruction module (Sec.~\ref{sec:CVR}) with text-to-vision reconstruction loss are utilized to obtain a comprehensive cross-modal understanding of the textual expression and visual context.
We elaborate on each component of our RecLMIS in detail in the following sections.

\begin{figure*}[ht]
\centerline{\includegraphics[width=\linewidth]{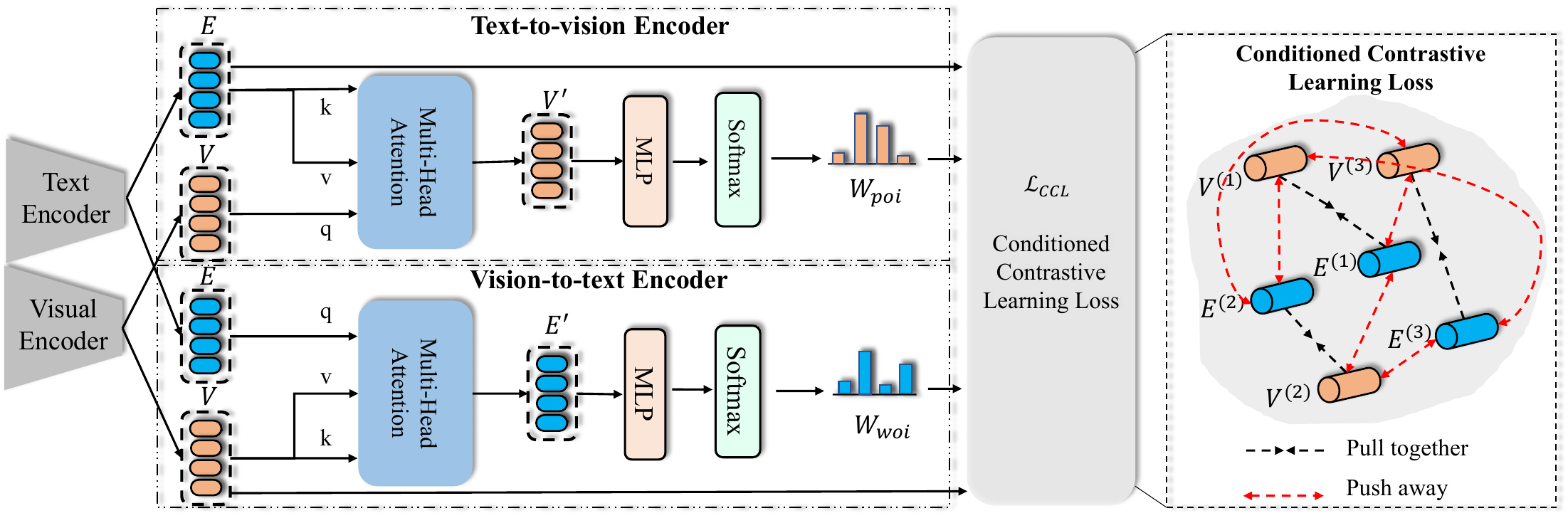}}
\caption{The structure of the Conditioned Interaction module.} \label{cim}
\end{figure*}

\subsection{Feature Encoder}

Given an input image $X$, we utilize the UNet~\cite{ronneberger2015u} backbone with the PLAM~\cite{li2023lvit} module as a visual encoder to extract visual features $V= \left\{v_{n}\right\}_{n=1}^{N}\in \mathbb{R}^{C \times H \times W}$,
where $C$, $H$, and $W$ denote the number of channels, height, and width, respectively. The $N$ is equal to $H  \times  W$.
Given a medical note \( T \), we use the BPE encoding algorithm to convert the sentence into a list of token IDs \( T' \in\mathbb{R}^{L} \), where $L$ is the number of tokens. Then, we use a word embedding layer to map the token ID list \( T' \) to high-dimensional vectors \( T'' \in\mathbb{R}^{L} \). Finally, we input \( T'' \) into a 12-layer standard transformer encoding block to extract the text feature representation vector $E= \left\{e_{l}\right\}_{l=1}^{L}\in\mathbb{R}^{C  \times  L}$.


\subsection{Conditioned Interaction Module}
\label{sec: ci}

In language-guided image segmentation tasks, the fusion of visual and textual features directly impacts the segmentation results. As shown in Fig.~\ref{fig: motivation}(a) in Sec.~\ref{sec:intro}, if the visual and textual features are not well aligned, the text prompt will fail to influence the segmentation results, contradicting the initial purpose of using text to guide segmentation. Numerous studies have demonstrated that using cross-attention to fuse features is currently the optimal choice, and combining it with contrastive learning loss can better align visual and textual features~\cite{hu2023beyond,chen2022multi,liu2023multi}. Although these methods have achieved notable results, there is still room for optimization in model and loss design. As shown in Fig.~\ref{fig:arch_compare}(b) and (c), these methods employ multiple fusion layers, resulting in significant parameters and computational overhead during the inference stage. Additionally, the contrastive learning losses in some studies~\cite{hu2023beyond,dawidowicz2023limitr,huang2021gloria} are not sufficiently fine-grained. Therefore, we aim to compress the fusion layers to a single layer and extend the contrastive loss to a more fine-grained patch-word level to reduce the impact of non-essential text tokens and image patches. 

\subsubsection{Hierarchical Structure}

As illustrated in Fig.\ref{cim}, the CI module is implemented in two parts: text-to-vision encoder and vision-to-text encoder. 
Compared to previous works (\eg, LViT~\cite{li2023lvit}), RecLMIS does not require complex multi-level interactive alignment of features to achieve good results because it explicitly models the semantic connections between language and vision by automatically learning to reconstruct key features in text and images during training, more details in Sec.~\ref{sec:CVR} and Sec.~\ref{sec:CLR}. Specifically, given the input text features $E$ and visual features $V$, we utilize cross-attention operation to perform cross-modal feature interaction. For the vision-to-text encoder, we regard $E$ as the query and $V$ as the key and value to obtain a new text feature $E'$. Here, the visual feature $V$ is first flattened to $\mathbb{R}^{C\times N}$. Then the feature interaction is expressed as:
\begin{equation}
E'=\mathrm{softmax}(\frac{(W_{q}E)^\top(W_{k}V)}{\sqrt{\hat{C}}})\left(W_{v}V\right)^\top,
\end{equation}
where $W_{q}$, $W_{k}$ and $W_{v}$ are the learnable parameters. 

\begin{figure*}[ht]
\centerline{\includegraphics[width=\linewidth]{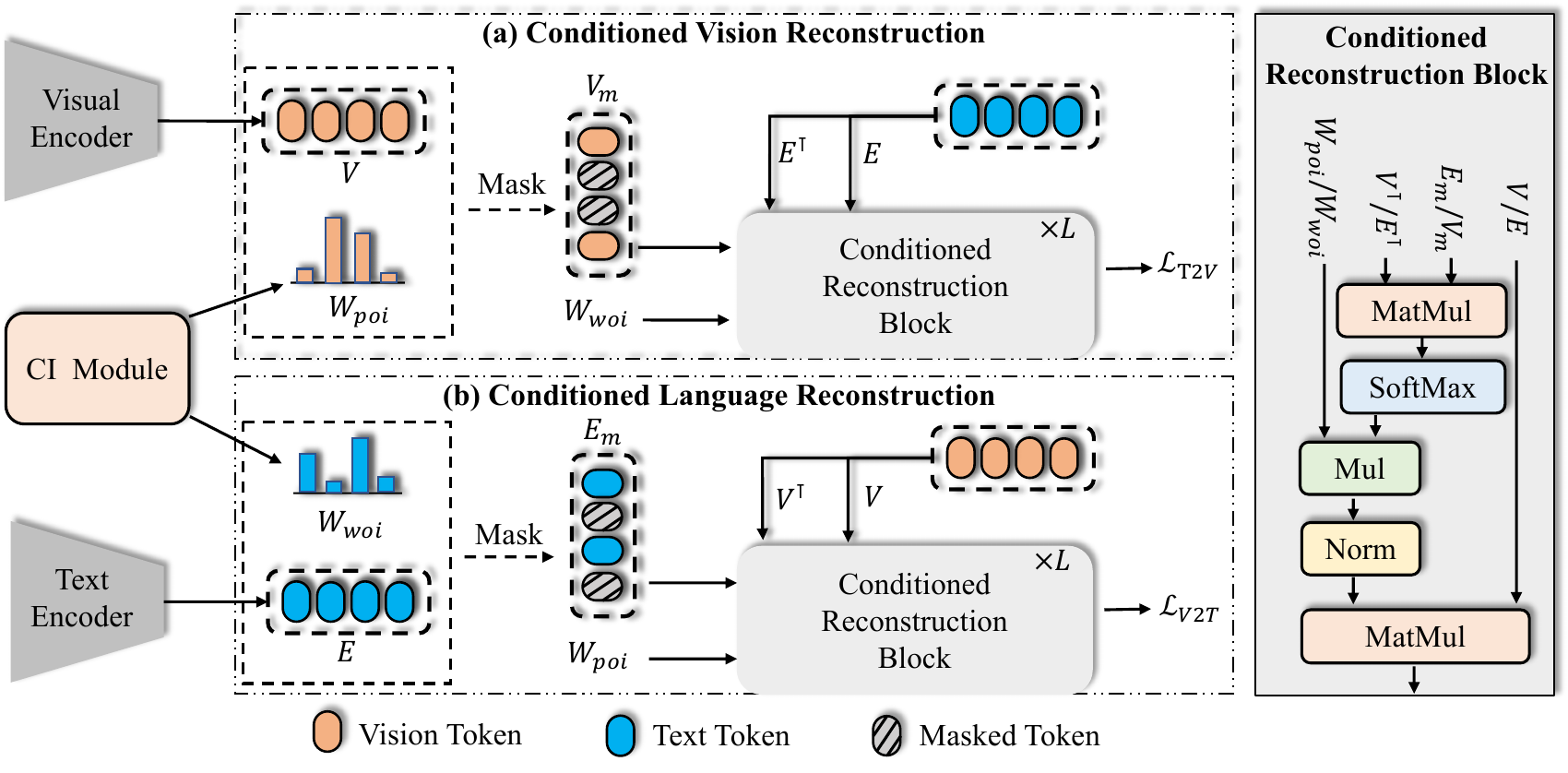}}
\caption{The structure of (a) Conditioned Vision Reconstruction module and (b) Conditioned Language Reconstruction.} \label{fig:CLR_CVR}
\end{figure*}


Similarly, we consider $V$ as the query and $E$ as the key and value in the text-to-vision encoder to get a new visual feature $V'$.
Thus, Utilizing the final layer of coarse-grained image features within the backbone architecture enables us to capture the relationship between a single word and a patch of the image.
More precisely, we can obtain the patches of interest $W_{poi}\!=\!\{w_v^{i}\}_{i=1}^{N}$ with shape $1 \times N$ and the word of interest $W_{woi}\!=\!\{w_e^{j}\}_{j=1}^{L}$ with shape $1 \times L$ by the following equation:
\begin{equation}
\label{eq:w_poi}
W_{poi}=\mathrm{softmax}(\mathrm{MLP}(V^{\prime})),
\end{equation}
\begin{equation}
\label{eq:w_woi}
W_{woi}=\mathrm{softmax}(\mathrm{MLP}(E^{\prime})).
\end{equation}

\subsubsection{Conditioned Contrastive Learning}
\label{ccl}

The vanilla contrastive learning~\cite{hu2023beyond,radford2021learning} pools all visual and textual features equally, which ignores the uncertainty in vision-text alignment. 
 The $\mathcal{A}_{agg}$ loss in Limitr~\cite{dawidowicz2023limitr} is more fine-grained. The $\mathcal{A}_{agg}$ primarily considers the contribution of each text token to the similarity calculation. Let the patch-word alignment matrix $A\!=\![a_{ij}]^{N\times L}$, 
where $a_{ij}\!=\! {\frac{v_i^\top e_j}{||v_i|| ||e_j||}}$
denotes the alignment score between $i\text{-th}$ patch and $j \text{-th}$ word. Thus, the similarity between an image-text pair in Limitr~\cite{dawidowicz2023limitr} can be expressed as:
\begin{equation}
\label{eq:similarity_socre}
\mathcal{S}_{v,e}=\sum_{j=1}^{L}w_e^ja_{ij},
\end{equation}
where $w_e^j$ denotes the attention weight of the $j$-th word. Although this method finely aligns image patches to each text token, it does not consider the relationship between each text token and every image patch in a fine-grained manner. Additionally, noisy regions in the text or image might interfere with the similarity calculation.

Inspired by fine-grained semantic alignment~\cite{yao2021filip}, we maintain the sequence structure of both the image and text.
We introduce $w_e^j$ to calculate the contribution of each image patch to the similarity calculation. Additionally, to mitigate the impact of noise, we select only the token/patch with the highest similarity to participate in the calculation. The similarity between an image-text pair in our $\mathcal{L}_{CCL}$ calculation formula can be expressed as follows:
\begin{equation}
\label{eq:similarity_socre}
\mathcal{S}_{v,e}=\frac12(\sum_{i=1}^{N}w_v^i\max_ja_{ij}+\sum_{j=1}^{L}w_e^j\max_ia_{ij}).
\end{equation}
Finally, the cross-modal contrastive loss~\cite{wang2022disentangled} can be expressed as:
\begin{equation}
\label{eq:L_ccl}
\begin{aligned}
\mathcal{L}_{\mathrm{CCL}}& =-\frac12[\frac1B\sum_{i=1}^{B}\log\frac{\exp\left(\mathcal{S}_{v_{i},e_{i}}\right)/\tau}{\sum_{j=1}^{B}\exp\left(\mathcal{S}_{v_{i},e_{j}}\right)/\tau}  \\
&+\frac{1}{B}\sum_{i=1}^{B}\log\frac{\exp\left(\mathcal{S}_{v_{i},e_{i}}\right)/\tau}{\sum_{j=1}^{B}\exp\left(\mathcal{S}_{v_{j},e_{i}}\right)/\tau}],
\end{aligned}
\end{equation}
where $\tau$ is the temperature hyperparameter and $B$ is the batch size.

\subsection{Conditioned Vision Reconstruction}
\label{sec:CVR}

Several studies have demonstrated the effectiveness of masking and reconstruction in enhancing representation learning~\cite{he2022masked,devlin2018bert,xie2022simmim,li2023masked,li2023mage,chen2022multi}, utilizing various reconstruction techniques or visual generation methods to improve model representational capacity and downstream task performance. These methods all fall under the category of Fig.~\ref{fig:arch_compare}(a), which involves masking and reconstructing the image itself. While such approaches can enhance the model's text-to-vision representation capability, reconstructing the original image includes unimportant regions, thus adding unnecessary computational complexity. We aim to introduce an auxiliary module that, without altering the original model structure, utilizes text features \(E\) to reconstruct masked key areas of the visual features \(V_m\), thereby achieving text-to-vision alignment.


\begin{table*}[ht]
\centering
\caption{Comparisons to current state-of-the-art methods on QaTa-COV19 and MosMedData+. ``$\Delta$'' denotes the difference between the results obtained using our RecLMIS and LViT~\cite{li2023lvit}. The highest scores are bolded and we compute the difference between the results obtained using our approach (RecLMIS) and LViT~\cite{li2023lvit}
}\label{sota_chest}
\resizebox{0.9\textwidth}{!}{%
\begin{tabular}{lcccccccc}
\toprule
\multirow{2}{*}{\textbf{Method}} & \multirow{2}{*}{\textbf{Backbone}} & \multirow{2}{*}{\textbf{Text}}  & \multirow{2}{*}{\textbf{Parameters (M)}}                       & \multirow{2}{*}{\textbf{FLOPs (G)}} & \multicolumn{2}{c}{\textbf{QaTa-COV19}} & \multicolumn{2}{c}{\textbf{MosMedData+}} \\
\cmidrule{6-7} \cmidrule{8-9}
& & & & &\textbf{Dice(\%)$\uparrow$}                     & \textbf{mIoU(\%)$\uparrow$} &\textbf{Dice(\%)$\uparrow$}                     & \textbf{mIoU(\%)$\uparrow$}                     \\ \midrule
U-Net~\cite{ronneberger2015u}                          & CNN & \XSolidBrush   & 14.8                                    & 50.3                                    & 79.02                                 & 69.46 & 64.60 & 50.73                                \\
U-Net++~\cite{zhou2018unet++}                        & CNN & \XSolidBrush   & 74.5                                    & 94.6                                    & 79.62                                 & 70.25  & 71.75 & 58.39                               \\
nnUNet~\cite{isensee2021nnu}                         & CNN & \XSolidBrush   & 19.1                                    & 412.7                                   & 80.42                                 & 70.81       & 72.59  & 60.36                         \\
TransUNet~\cite{chen2021transunet}                  & Hybrid & \XSolidBrush   & 105.0                                   & 56.7                                    & 78.63                                 & 69.13         & 71.24 & 58.44                        \\
Swin-Unet~\cite{cao2022swin}                   & Transformer & \XSolidBrush   & 82.3                                    & 67.3                                    & 78.07                                 & 68.34              & 63.29 &  50.19                  \\
\midrule
ConVIRT~\cite{zhang2022contrastive}                       & CNN & \CheckmarkBold & 35.2                                    & 44.6                                    & 79.72                                 & 70.58      & 72.06 & 59.73                           \\
TGANet~\cite{tomar2022tganet}                        & CNN & \CheckmarkBold & 19.8                                    & 41.9                                    & 79.87                                 & 70.75           & 71.81 &  59.28                     \\
CLIP~\cite{radford2021learning}                      & Transformer & \CheckmarkBold & 87.0                                    & 105.3                                   & 79.81                                 & 70.66           & 71.97 &  59.64                     \\
GLoRIA~\cite{huang2021gloria}                     & CNN & \CheckmarkBold & 45.6                                    & 60.8                                    & 79.94                                 & 70.68               & 72.42 & 60.18                  \\
ViLT~\cite{kim2021vilt}                       & Hybrid & \CheckmarkBold & 87.4                                    & 55.9                                    & 79.63                                 & 70.12                  & 72.36 &  60.15              \\
LAVT~\cite{yang2022lavt}                        & Transformer & \CheckmarkBold & 118.6                                   & 83.8                                    & 79.28                                 & 69.89                & 73.29 &  60.41                \\
LViT~\cite{li2023lvit}                       & Hybrid & \CheckmarkBold & 29.7                                    & 54.1                                    & 83.66                                 & 75.11                    & 74.57 & 61.33             \\ 
SLViT~\cite{ouyang2023slvit}                       & Hybrid & \CheckmarkBold & 131.5                                    & 51.1                                    & 79.25                                 & 68.87                    & 72.57 & 60.78             \\ 
DMMI~\cite{hu2023beyond}                       & Transformer & \CheckmarkBold & 114.6                                    & 63.3                                    & 84.13                                 & 75.66                   & 75.01 & 61.83             \\ 
RefSegformer~\cite{wu2024towards}                       & Transformer & \CheckmarkBold & 195.0                                    & 103.6                                    & 84.09                                 & 75.48                    & 74.98 & 61.70             \\ \midrule

\textbf{RecLMIS}             & CNN & \CheckmarkBold & 23.7                           & \textbf{24.1}                           & \textbf{85.22}                        & \textbf{77.00}           & \textbf{77.48} & \textbf{65.07}             
\\ \bottomrule
\end{tabular}
}
\end{table*}

As shown in Fig.~\ref{fig:CLR_CVR}(a), we consider $W_{poi}$ obtained by the CI module as the masking probability distribution. In other words, the visual feature $v_i$ with higher $w_v^i$ has a higher masking probability. Define the mask ratio as $\alpha_{v} \in [0,1]$, we then set the $m\!=\! \alpha_{v}\! \cdot\! N$ patch features to zero to generate the masked visual feature $V_{m}$.
Specifically, we pinpoint the patches of most significance to the model by utilizing the $W_{poi}$ obtained from the CI module through a probability distribution approach. 
We then modify the visual features by setting the features corresponding to these positions to zero, ultimately producing the masked visual features $V_{m}$.
Ideally, a model with a comprehensive understanding of image and text semantics is able to reconstruct $V_{m}$ well based on complete text features $E$ and words of interest $W_{woi}$. Inspired by~\cite{zheng2022weaklya, zheng2022weaklyb}, we introduce a conditioned reconstructor, denoted as $\Psi$, to assess the semantic alignment between the words of interest and the patches of interest. As shown in Fig.~\ref{model}, our conditioned reconstructor $\Psi$ takes $W_{woi}$ as a condition, masked visual features $V_{m}$ as the query, and the textual feature $E$
as the key and value. We employ the dot-product attention mechanism to capture cross-modal interactions. The reconstructed vision features $\hat{V} = \{\hat{v}\}_{i=1}^{N}$ can be formulated as:
\begin{equation}
\begin{aligned}
\hat{V}&=\Psi(V_m,E,W_{woi})
\\
&= \operatorname{softmax}\left(\frac{V_m E^\top}{\sqrt{D}}\odot W_{woi}\right)E,
\end{aligned}
\end{equation}
where $\sqrt{D}$ is the feature dimension and $\odot$ denotes the Hadamard product. 

Then we utilize the mean squared error to measure the similarity between the reconstructed $\hat{V}$ and the original visual features $V$:
\begin{equation}
\mathcal{L}_{\mathrm{T} 2 \mathrm{V}}=\frac{1}{B} \sum_{i=1}^{B} \frac{1}{N} \sum_{j=1}^{N} w_{v}^{i j}\left|v_{i j}-\hat{v}_{i j}\right|^{2}.
\end{equation}

\subsection{Conditioned Language Reconstruction}
\label{sec:CLR}
Using masked text to reconstruct raw text has been well-developed in method BERT~\cite{devlin2018bert}. Many works~\cite{chen2022multi,hu2023beyond} have adopted such approaches to enhance image-to-text representation capabilities. These methods fall under the category of Fig.~\ref{fig:arch_compare}(a), which involves masking and reconstructing the text itself. While such schemes can improve the model's vision-to-text representation capabilities, reconstructing the original text includes redundant words, introducing noise. We aim to introduce an auxiliary module that, without modifying the original model structure, utilizes visual features \(V\) to reconstruct the masked keywords in the text features \(E_m\), thereby achieving a better understanding from vision to text.

As shown in Fig.~\ref{fig:CLR_CVR}(b), to reconstruct masked textual features from visual features, we perform a similar masking operation as described in CVR~\ref{sec:CVR}. Let's define the reconstructed text features as $\hat{E} = \{\hat{e}\}_{i=1}^{L}$ and $E_m$ denotes the masked textual features. The process of conditioned language reconstruction can be expressed as:
\begin{equation}
\begin{aligned}
\hat{E}&=\Psi(E_m,V,W_{poi})
\\
&= \operatorname{softmax}\left(\frac{E_m V^\top}{\sqrt{D}}\odot W_{poi}\right)V.
\end{aligned}
\end{equation}
The similarity between the reconstructed $\hat{E}$ and the textual features $E$ can be similarly computed by mean squared error, which is as follows:
\begin{equation}
\mathcal{L}_{\mathrm{V} 2 \mathrm{T}}=\frac{1}{B} \sum_{i=1}^{B} \frac{1}{L} \sum_{j=1}^{L} w_{e}^{i j}\left|e_{i j}-\hat{e}_{i j}\right|^{2}.
\end{equation}

\begin{table*}[h]
\caption{The efficiency of the different models on the QaTa-COV19 dataset. 'Cov.' and 'Inf.' denotes 'Convergence' and 'Inference', respectively.}
\label{tab:efficiency}
\centering
\begin{tabular}{cccc|ccc}
\toprule
\multirow{2}{*}{Method} & \multicolumn{3}{c|}{\textbf{Training Phase}}                      & \multicolumn{3}{c}{\textbf{Inference Phase}}                          \\ \cmidrule{2-7} 
                        & \textbf{Params (M)} & \textbf{FLOPs (G)} & \textbf{Con. Time (hours)} & \textbf{Params (M)} & \textbf{FLOPs (G)} & \textbf{Inf. Time (ms/img)} \\ \midrule
LAVT~\cite{yang2022lavt}                    & 118.6               & 83.8               & -                      & 118.6               & 83.8              & 170.3                       \\
LViT~\cite{li2023lvit}                    & \textbf{29.7}                & 54.1               & 16                    & 29.7                & 54.1              & 37.8                        \\ 
DMMI~\cite{hu2023beyond}                    & 114.6                & 63.3               & 8.6                    & 114.6                & 63.3              & 67.4                        \\ 
RefSegformer~\cite{wu2024towards}                    & 195.0                & 103.6               & 13.5                    & 195.0                & 103.6               & 75.6                        \\ \midrule
\textbf{RecLMIS(ours)}  & 74.2                & \textbf{36.62}     & \textbf{7.5}          & \textbf{23.7}       & \textbf{24.1}     & \textbf{20.3}               \\ \bottomrule
\end{tabular}%
\end{table*}

\subsection{Training and Inference}

\textbf{Training.} To predict accurate segmentation results, we introduce the dice loss $\mathcal{L}_{\mathrm{Dice}}$ and cross-entropy loss $\mathcal{L}_{\mathrm{CE}}$ commonly used in image segmentation~\cite{chen2021transunet, huang2020unet,li2023lvit}. Finally, the full objective of semantic alignment can be formulated as: 

\begin{equation}
\mathcal{L}\!=\! \lambda_1 \mathcal{L}_{\mathrm{V} 2 \mathrm{T}} + \lambda_2 \mathcal{L}_{\mathrm{T} 2 \mathrm{V}}+ \lambda_3 \mathcal{L}_{\mathrm{CCL}} + \lambda_4(\mathcal{L}_{\mathrm{Dice}} + \mathcal{L}_{\mathrm{CE}}),
\end{equation}
where $\lambda_1$, $\lambda_2$, $\lambda_3$ and $\lambda_4$ are trade-off hyper-parameters.

\textbf{Inference.} 
As shown in Fig.~\ref{model}, the orange arrows denote the inferring path. We only need to keep the visual encoder, visual decoder, and text-to-vision encoder in the CI module.
In contrast to previous work (\eg, \cite{li2023lvit}, \cite{yang2022lavt}), our RecLMIS eliminates the need for complex multi-level interaction, which significantly accelerates inference, as detailed in Sec.\ref{sec: efficiency}).

\section{Experiments}

\subsection{Datasets and Metrics}
\label{sec: data}

To evaluate the effectiveness of our method, we conducted experiments on two medical datasets. 

\subsubsection{QaTa-COV19 Dataset}
QaTa-COV19~\cite{degerli2022osegnet}, which consists of 9258 COVID-19 chest X-ray radiographs with extended medical notes by~\cite{li2023lvit}. The data split follows LViT~\cite{li2023lvit}, \ie, the number of images in the training set, validation set, and test set are 5716, 1429, and 2113, respectively.

\subsubsection{MosMedData+ Dataset}
MosMedData+~\cite{morozov2020mosmeddata, hofmanninger2020automatic, li2023lvit}, which contains 2729 CT scan slices of lung infections. The data split following LViT~\cite{li2023lvit}, the number of images in the training set, validation set, and test set are 2183, 273, and 273, respectively.

\subsubsection{Metrics}
For the evaluation metrics, the Dice score and the mIoU metric are used to evaluate performance, which can be expressed as:
\begin{equation}
Dice=\sum_{i=1}^{\hat{N}}\sum_{j=1}^{\hat{C}}\frac{1}{{\hat{N}}{\hat{C}}}\cdot\frac{2|p_{ij}\cap y_{ij}|}{(|p_{ij}|+|y_{ij}|)},
\end{equation}
\begin{equation}
    m I o U=\sum_{i=1}^{\hat{N}} \sum_{j=1}^{\hat{C}} \frac{1}{\hat{N} {\hat{C}}} \cdot \frac{\left|p_{i j} \cap y_{i j}\right|}{\left|p_{i j} \cup y_{i j}\right|},
\end{equation}
where $\hat{N}$ represents the number of pixels, ${\hat{C}}$ represents the number of categories, $p_{ij}$ represents the prediction class that pixel $i$ belongs to category $j$, and $y_{ij}$ represents whether pixel $i$ belongs to category $j$.

\begin{table}[t]
\centering
\caption{Main ablation study of proposed components on QaTa-COV19 dataset.}
\label{table: components}
\resizebox{0.9\columnwidth}{!}{%
\begin{tabular}{c|ccccc}
\toprule
&\textbf{CI}   & \textbf{CLR}   & \textbf{CVR}   & \textbf{Dice(\%)$\uparrow$}  & \textbf{mIoU(\%)$\uparrow$}  \\ \midrule
(a)&\CheckmarkBold & \CheckmarkBold & \CheckmarkBold & \textbf{84.84}     & \textbf{76.44}    \\
(b)&\CheckmarkBold & \XSolidBrush   & \CheckmarkBold & 84.43              & 75.98    \\
(c)&\CheckmarkBold & \CheckmarkBold & \XSolidBrush   & 84.30              & 75.84             \\
(d)&\CheckmarkBold & \XSolidBrush   &  \XSolidBrush   & 82.52              & 73.35             \\
(e)&\XSolidBrush     & \XSolidBrush     & \XSolidBrush  & 79.37              & 70.00             \\ \bottomrule

\end{tabular}
}
\end{table}

\begin{table}[t]
\centering
\caption{Ablation study concerning the Inclusion of condition on QaTa-COV19 dataset. ``Con" means the condition in the current module exists.}
\label{table: condition}
\resizebox{\columnwidth}{!}{%
\begin{tabular}{c|c|c|cc}
\toprule
\textbf{CCL}   & \textbf{CLR}   & \textbf{CVR}   & \textbf{Dice(\%)$\uparrow$}  & \textbf{mIoU(\%)$\uparrow$}  \\ \midrule
               &                &                & 84.14              & 75.57             \\
Condition            &                &                & 84.56              & 75.97    \\
               & Condition &               & 84.35              & 75.94        \\
                &                & Condition & 84.63             & 76.19        \\
Condition & Condition & Condition & \textbf{84.84}     & \textbf{76.44}    \\ 
\bottomrule
\end{tabular}
}
\end{table}

\subsection{Implementation Details}
We implemented our method in PyTorch~\cite{paszke2019pytorch} and the training devices were four NVIDIA GeForce RTX 3090 with 24GB video memory capacity per card. The initial learning rate was set to 3e-4, and we utilized cosine annealing as the scheduler. The default batch size was 32 and the resolution of the image was $224\! \times \!224$. We trained the networks on GPUs with 200 epochs for QaTa-COV19~\cite{degerli2022osegnet} dataset and 500 epochs for MosMedData+~\cite{morozov2020mosmeddata, hofmanninger2020automatic, li2023lvit}. The visual mask ratio $\alpha_{v}$ and textual mask ratio $\alpha_{t}$ were 0.5 and 0.3, respectively. The number of conditioned transformer layers was three. Besides, we applied Gaussian blur with random sigma, contrast distortion, saturation distortion, and hue distortion to augment the datasets. 
The $\lambda_1, \lambda_2, \lambda_3, \lambda_4$ were 1, 1, 0.2, and 5, respectively. 

\subsection{Comparison with State-Of-The-Art}
\label{sec: sota}
We compare our network against several mainstream CNN and Transformer based segmentation models. Furthermore, we also provide the number of network parameters and the computational cost of different methods.

\subsubsection{Comparison on QaTa-COV19 and MosMedData+ Dataset} 

\begin{figure*}[ht]
\centerline{\includegraphics[width=\linewidth]{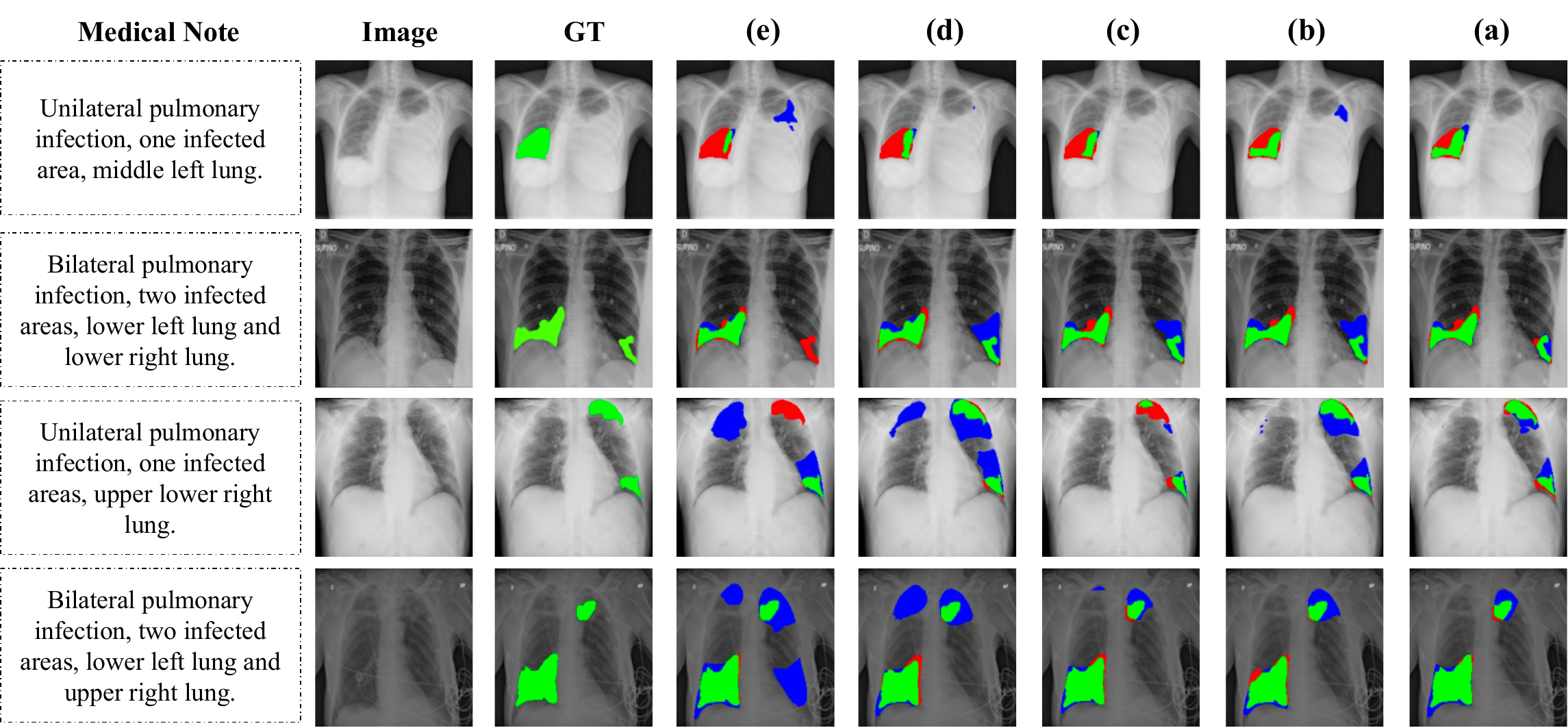}}
\caption{The visualization of the main ablation experiments Sec.~\ref{sec:Effect of Proposed Components}). The sequence numbers correspond to those in Table~\ref{table: components}. Green, red, and blue, indicate true positive, false negative, and false positive pixels, respectively.} \label{fig:vis_ablation}
\end{figure*}

\begin{figure}[t]
\centerline{\includegraphics[width=\linewidth]{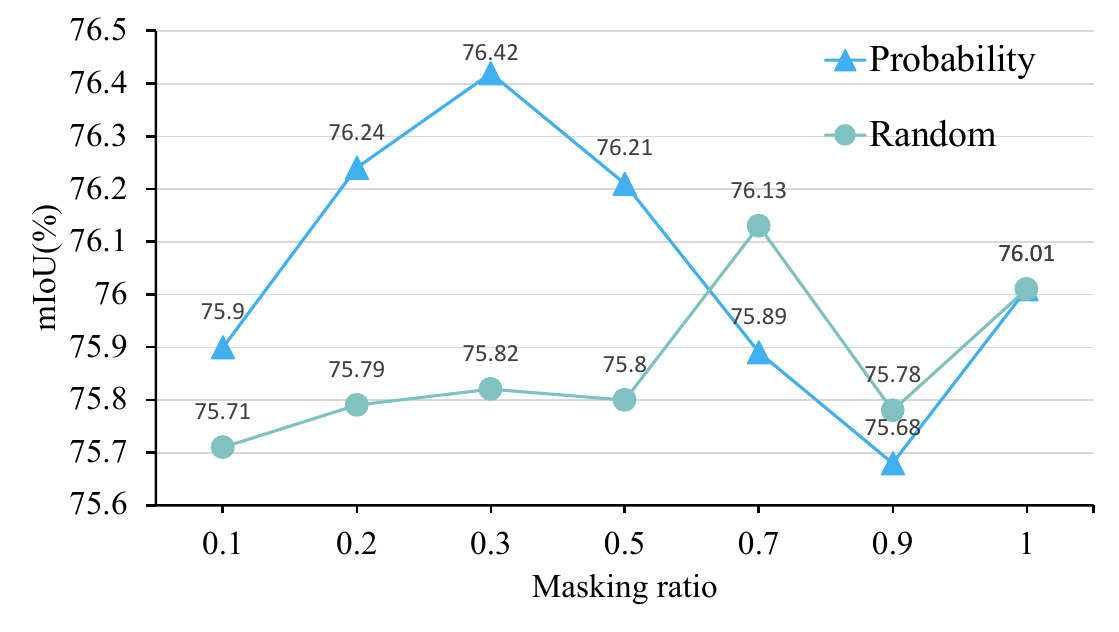}}
\caption{Effect of the masking strategy and ratio on QaTa-COV19 dataset.} \label{ablation: mask_stategy}
\end{figure}

\begin{figure}[t]
\centerline{\includegraphics[width=\linewidth]{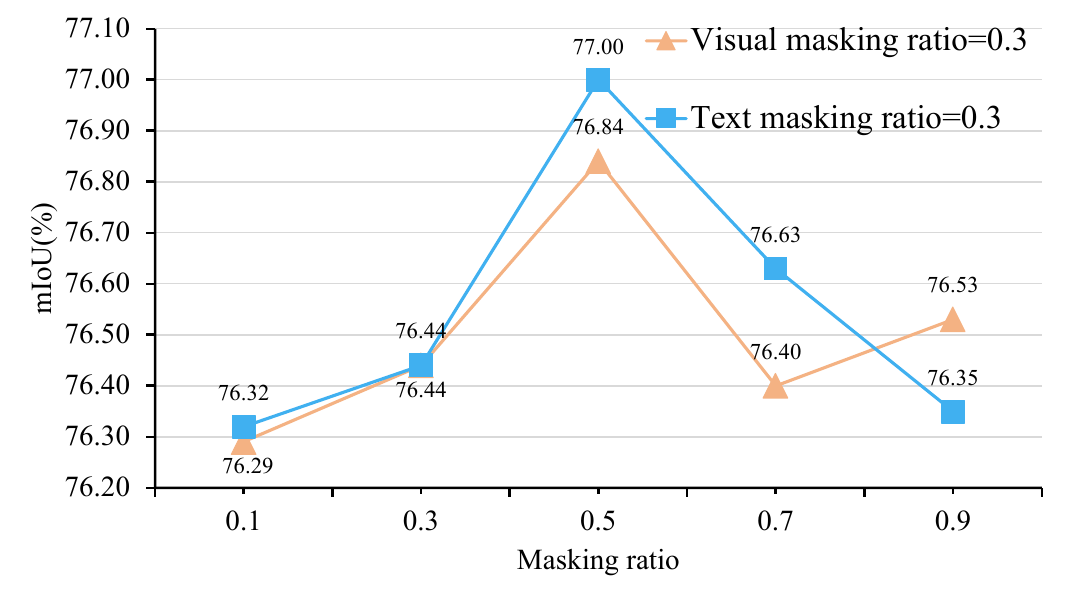}}
\caption{Effect of the masking strategy and ratio on QaTa-COV19 dataset.} \label{ablation: mask_ratio}
\end{figure}

In Table~\ref{sota_chest}, we evaluate RecLMIS against the other segmentation methods on the QaTa-COV19~\cite{degerli2022osegnet} and MosMedData+~\cite{morozov2020mosmeddata}.
Compared to the previous optimal model LViT~\cite{li2023lvit}, our rec surpasses it by 1.33 mIoU and 1.18 Dice on QaTa-COV19 and 2.91 mIoU and 3.74 Dice on MosMedData+.
More encouragingly, RecLMIS exhibits a relative reduction of 55.5\% in computational cost and a relative decrease of 20.2\% in parameter count. These results demonstrate the superiority and efficiency of RecLMIS.
Moreover, RecLMIS improves the Dice score by 4.42 compared to the suboptimal nnUNet~\cite{isensee2021nnu} without text on QaTa-COV19~\cite{degerli2022osegnet}, which indicates that introducing medical notes can improve model performance effectively in chest X-ray radiograph segmentation.

\begin{figure*}[t]
\centerline{\includegraphics[width=\textwidth]{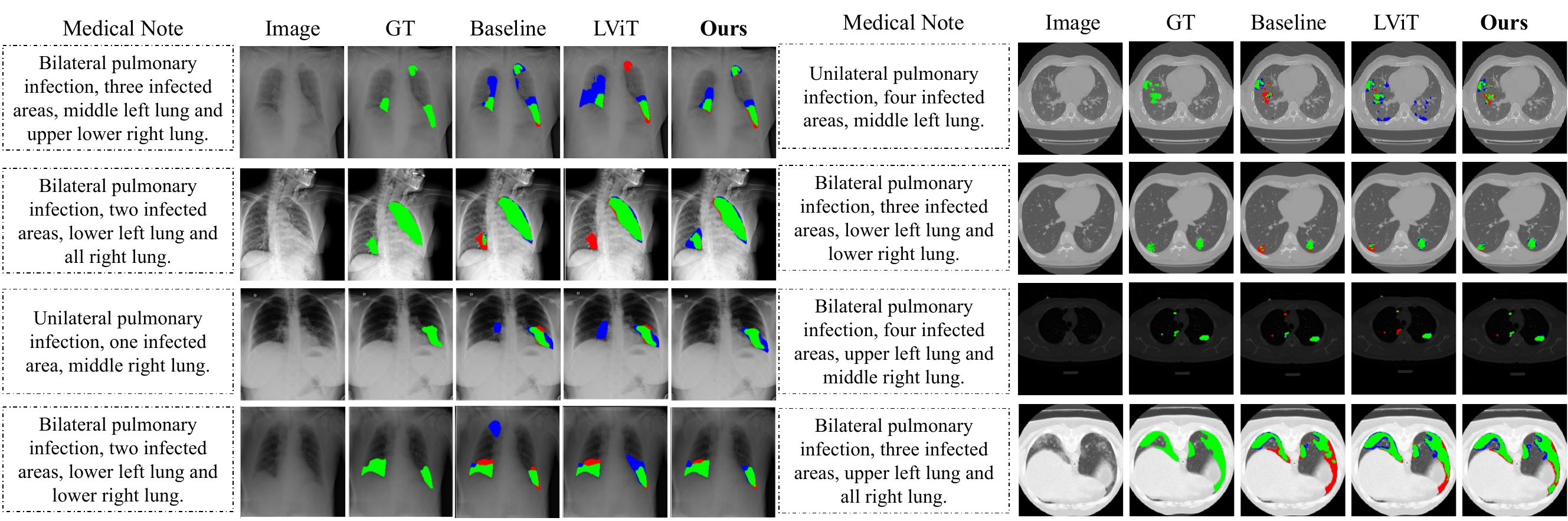}}
\caption{The visualization of the main comparison. The left-hand image represents the visualization results on the test set of QaTa-COV19, while the right-hand image represents the visualization results on the test set of MosMedData+. Green, red, and blue, indicate true positive, false negative, and false positive pixels, respectively. Orange font signifies words that carry instructive significance for the segmentation results. Best viewed in color, and further enhance the view by zooming in.} \label{vis}
\end{figure*}

As shown in Fig.~\ref{vis}, we conducted a visualization analysis of the main ablation experiments. The RecLMIS we proposed can accurately segment, regardless of size or shape. Whether on the QaTa-COV19~\cite{degerli2022osegnet}, MosMedData+~\cite{morozov2020mosmeddata, hofmanninger2020automatic, li2023lvit}, our model has outperformed LViT~\cite{li2023lvit}. 
\subsubsection{Efficiency comparison}
\label{sec: efficiency}

In Table~\ref{tab:efficiency}, we compute the inference time using an NVIDIA GeForce RTX 3090 on the QaTa-COV19~\cite{degerli2022osegnet} dataset. 
During the training phase, our RecLMIS demonstrates balanced performance. Despite a higher parameter count in RecLMIS training than other methods, both the FLOPs and training convergence time were significantly curtailed, resting at a mere 7.5 hours, less than half of the 16 hours required by LViT~\cite{li2023lvit}.
Upon entering the inference phase, RecLMIS continued to showcase superiority. Since we have discarded complex multi-level cross-modal interaction methods (\eg, LViT~\cite{li2023lvit}, LAVT~\cite{yang2022lavt}), as well as CVR and CLR can be removed during inference, our method achieves faster processing speeds during the inference phrase. Our model significantly reduced both the parameter count and FLOPs to 23.7M and 24.1G respectively, representing a reduction of 20.2\% and 55.5\% compared to LViT~\cite{li2023lvit}. Moreover, the inference time of RecLMIS was clocked at 20.3 milliseconds per image, a speed increase of 46.3\% compared to LViT~\cite{li2023lvit}, markedly outpacing both LAVT~\cite{yang2022lavt} and other methods~\cite{hu2023beyond, wu2024towards}.

\begin{figure*}[t]
\centerline{\includegraphics[width=\textwidth]{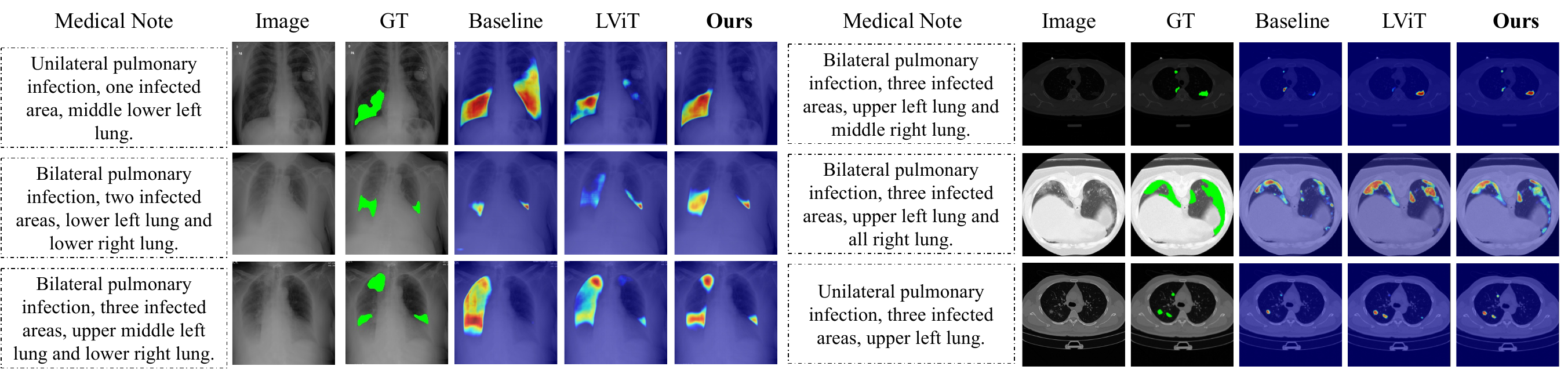}}
\caption{The visualization of the class activation heatmaps. The left-hand image represents the visualization results on the test set of QaTa-COV19, while the right-hand image represents the visualization results on the test set of MosMedData+.
The column titled ``medical note" denotes the input textual prompt, while the column labeled ``Image" signifies the input image. The column labeled ``GT" represents the ground truth segmentation target. Columns labeled ``Baseline", ``LViT" and ``Ours" depict the class activation heatmaps computed by these three models based on the gradient of the first layer of the decoder elucidating the efficacy of attention in the models. Best viewed in color, and further enhance the view by zooming in.} \label{cam}
\end{figure*}
\label{sec: vis}

In summary, RecLMIS not only outperforms other methods in mIoU metrics but also significantly curtails computational resource consumption and inference time, thus underscoring its potential and advantages in practical applications.

\subsection{Ablation Study}
\label{sec: ablation}
Ablation experiments on the challenging QaTa-COV19~\cite{degerli2022osegnet} dataset are conducted to substantiate the effectiveness of RecLMIS, which is explored in the following seven aspects. 
For all ablation experiments except those figures (Fig~\ref{ablation: mask_stategy}, Fig~\ref{ablation: mask_ratio}, Fig~\ref{fig:lamda_lr}), the text masking ratio and the visual masking ratio are both 0.3 in this section.

\subsubsection{Effect of Proposed Components}
\label{sec:Effect of Proposed Components}

We consider the U-Net~\cite{ronneberger2015u} model with the PLAM~\cite{li2023lvit} module but without textual prompts as our baseline. As shown in the last two rows of Table~\ref{table: components}, mIoU is significantly improved when we introduce the CI module, thanks to the textual information that provides effective guidance for segmentation. By comparing the first three rows, the inclusion of CLR and CVR results in an improvement of 0.3 and 0.8 absolute points in mIoU, respectively. This indicates that CVR and CLR play a significant role in facilitating effective alignment between the visual and textual modalities. Our full model achieves the best performance and outperforms the baseline model by 6.44 mIoU.

As shown in Fig.~\ref{fig:vis_ablation}, we conduct a visualization analysis of the main ablation experiments. It is evident that without the CI module incorporating textual information, there are many instances of incorrect positional predictions. When the CVR and CLR modules are introduced, the accurate boundary predictions are further refined.


\subsubsection{Effect of Condition}
As described in Sec.~\ref{sec: ci}, Sec.~\ref{sec:CVR}, and Sec.~\ref{sec:CLR}, we emphasize the interdependence and influence between contrastive learning and reconstruction processes in both textual and visual features. Therefore, in this section, 
we explore the effects introduced by condition constraints.
As shown in Table~\ref{table: condition}, the mIoU metric decreases when we remove the condition constraints $w_v$ and $w_e$ in any module. When we remove all conditioned constraints, the performance drops by 0.87 mIoU.

\subsubsection{Ablation Study on Mask Strategy}
\label{sec:mask_strategy}

We explore two masking mechanisms: one involves random text token and image patch selection, while the other relies on the obtained $W_{woi}$ and $W_{poi}$ as probability distributions to generate masks as Sec.~\ref{sec:CVR} mentioned. Employing the probability distribution-based method enables the model to autonomously learn the critical information locations in both images and text, which is more conducive to cross-modal learning.

\subsubsection{Effect of Masking Ratio}
As Sec.~\ref{sec:CVR} described, the probability distribution-based method works better, so we explore the impact of different text and image masking ratios built upon this foundation. Specifically, in Fig.~\ref{ablation: mask_ratio}, it can be observed that when the masking ratios for $\alpha_{t}$ and $\alpha_{v}$ are set to 0.3 and 0.5 respectively, the probability-based approach yields the best results. This reveals that images contain considerable redundant information, leading to better outcomes with larger masking ratios, consistent with the conclusions drawn in MAE. However, unlike MAE, our masking is not randomly selected but learned through the model. Hence, setting the image masking ratio to 0.9 in our experiments yields suboptimal results.

\subsubsection{Effect of the mechanism of cross-modal interaction}

\begin{table}[t]
\centering
\caption{Effect of the mechanism of cross-modal interaction on QaTa-COV19 dataset. ``Rec" includes CVR and CLR. ``CI" is the Conditioned Interaction module.}
\label{ablation: interact_mechanism}
\resizebox{\columnwidth}{!}{%
\begin{tabular}{c|c|cccc}
\toprule
\textbf{CI}     &\textbf{CVR\&CLR}     & \textbf{Dice(\%)$\uparrow$}  & \textbf{mIoU(\%)$\uparrow$}  \\ \midrule
Self-Attn              &Cross-Attn                 & 79.37                         & 69.87             \\
Self-Attn              &Self-Attn                 & 78.91                         & 69.51             \\
Cross-Attn              &Self-Attn                 & 84.50                         & 75.98             \\
Cross-Attn              &Cross-Attn                 & \textbf{84.84}                & \textbf{76.44}    \\ \bottomrule
\end{tabular}
}
\end{table}

Previous research has often utilized self-attention for aligning features across different modalities. Consequently, we investigate the effects of self-attention and cross-attention on model performance. To implement self-attention (SA) for reconstruction, we concatenate the original visual and text features as $k$ and $v$ and use the masked visual and text features concatenated as $q$ for self-attention computation. As depicted in Table~\ref{ablation: interact_mechanism}, it is evident that the simultaneous use of cross-attention (CA) in both Conditioned Interaction (CI) and reconstruction (CVR and CLR) represents the most optimal configuration.

\begin{figure}[t]
\centerline{\includegraphics[width=\linewidth]{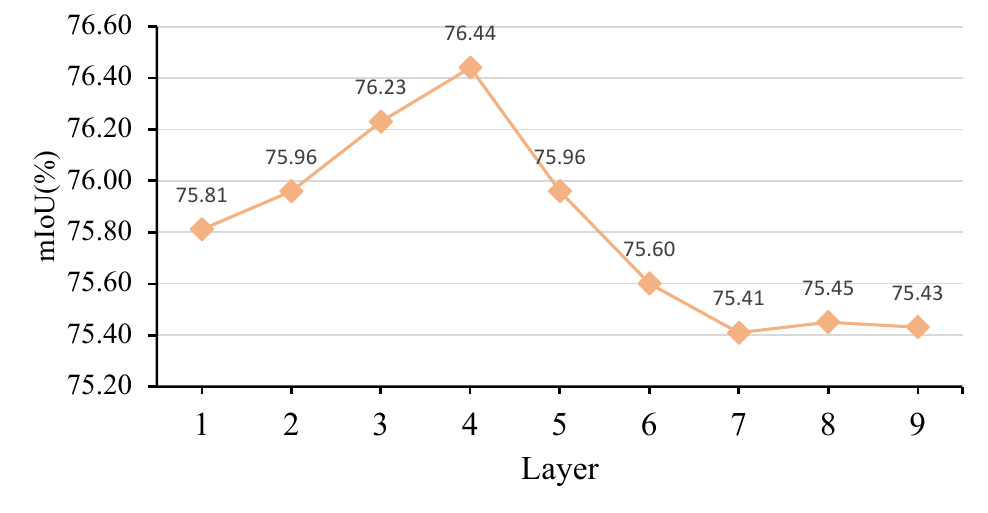}}
\caption{Effect of the number of reconstruction laryer on QaTa-COV19 dataset.} \label{ablation: rec_layer}
\end{figure}

\subsubsection{Effect of the Number of Reconstruction Layer}

As shown in Fig.~\ref{ablation: rec_layer}, we have explored the impact of varying the number of conditional reconstruction layers from 1 to 10 on performance. The experiments indicate that the best performance is achieved when the number of layers, L, is set to 3. We have observed that this is because of the limited volume of medical image data. With a larger number of layers (\eg, 7 layers), the reconstruction loss quickly decreases to near 0 during training, which can lead to overfitting. Conversely, with fewer than 3 layers (\eg, 1 layer), the reconstruction loss decreases slower, potentially resulting in underfitting.

\subsubsection{Effect of loss function weights and learning rate}
\begin{figure}[h]
\centerline{\includegraphics[width=\linewidth]{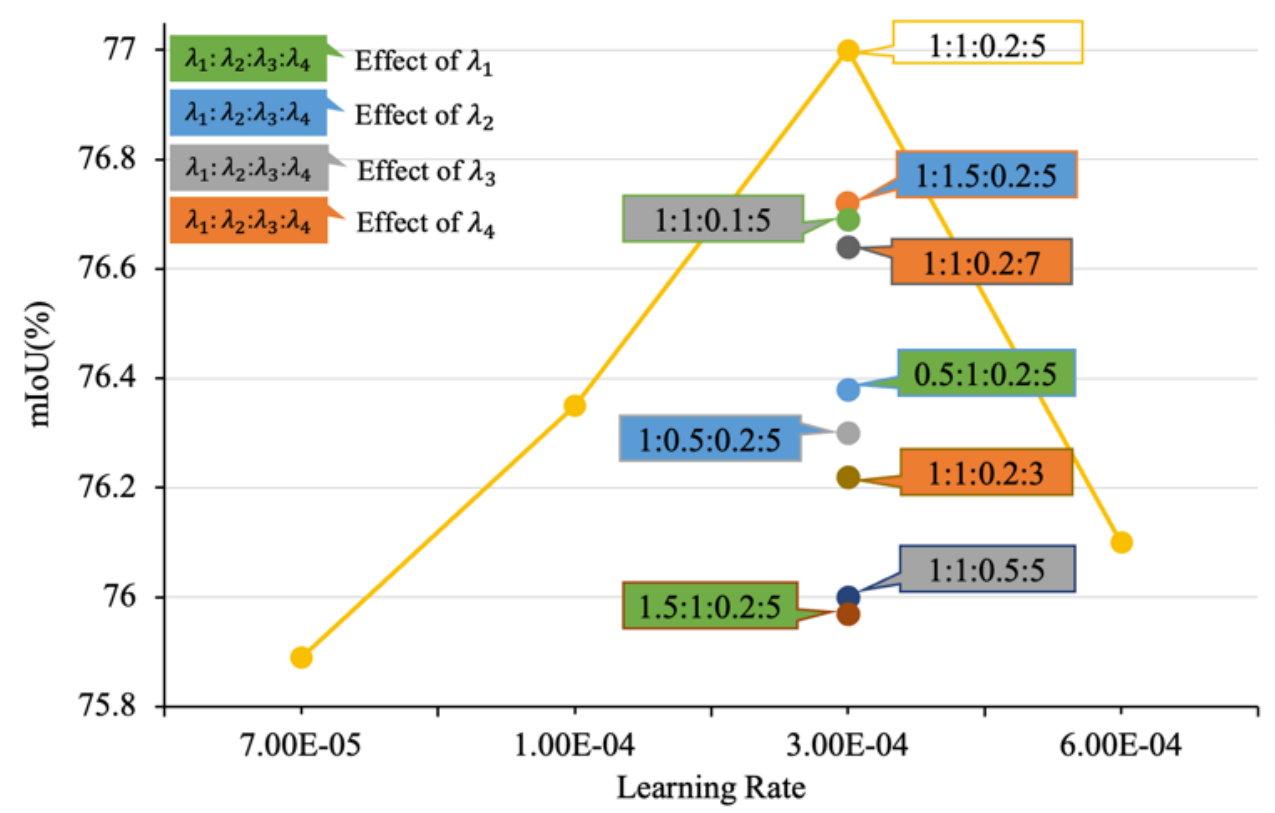}}
\caption{Effect of $\lambda_1$, $\lambda_2$, $\lambda_3$, $\lambda_4$ and the learning rate. Different background colors represent the impact brought by the variation of specific hyperparameters. the text masking ratio is 0.3 and the visual masking ratio is 0.5.} \label{fig:lamda_lr}
\end{figure}
We conducted ablation studies targeting $\lambda_1$, $\lambda_2$, $\lambda_3$, $\lambda_4$ and the learning rate in Fig.~\ref{fig:lamda_lr}. We observed the most significant performance decline when $\lambda_1$ and $\lambda_3$ were increased to 1.5 and 0.5, respectively. We believe this is due to the large auxiliary losses severely disrupting the convergence of the primary segmentation loss. Overall, despite these hyperparameter fluctuations, the minimum performance remained above LViT's 75.11.

\begin{figure*}[ht]
\centerline{\includegraphics[width=\linewidth]{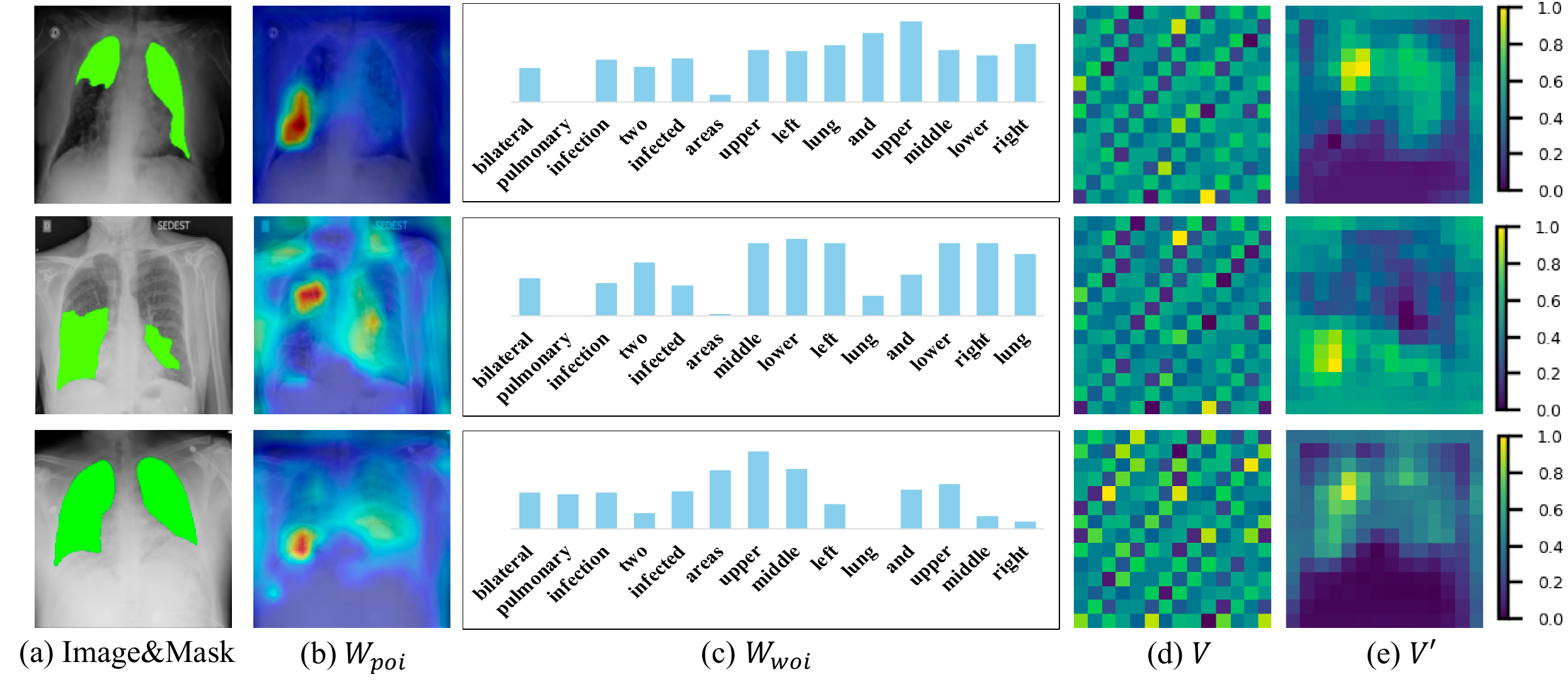}}
\caption{The visualization of key variables ($W_{poi},W_{woi},V$, and $V'$).} \label{fig:vis_feat}
\end{figure*}

\begin{figure*}[ht]
\centerline{\includegraphics[width=\linewidth]{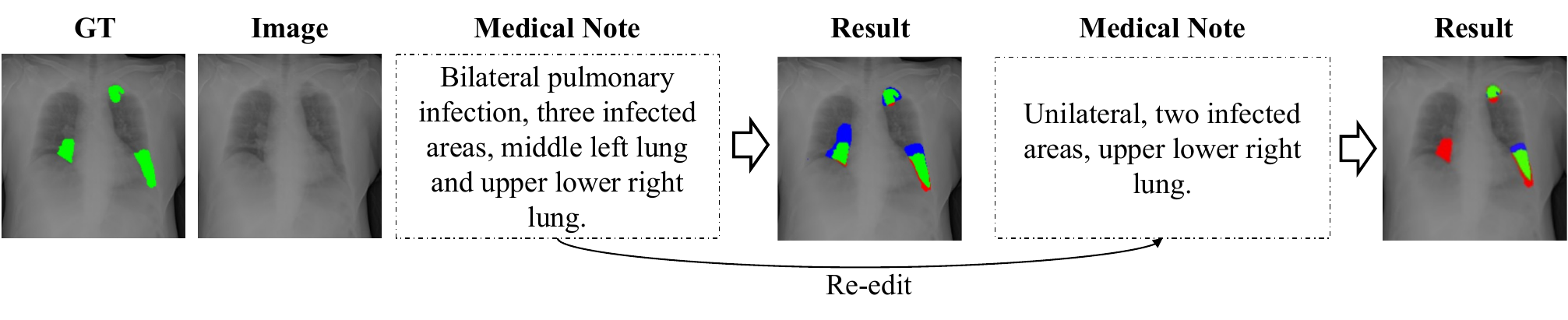}}
\caption{The visualization of the impact of text prompts on the results. Green, red, and blue, indicate true positive, false negative, and false positive pixels, respectively.} \label{fig:chage_text}
\end{figure*}

\section{Qualitative Analysis}


\textbf{Quantitative analysis of the model comparison.} For interpretability, we employ Grad-CAM~\cite{selvaraju2017grad} to illustrate the class activation heatmaps computed by these three models based on the gradient of the first layer of the decoder. 
As shown in Fig.~\ref{cam}, we find that RecLMIS can achieve improved alignment between visual and textual features, compared to LViT~\cite{li2023lvit}, RecLMIS focuses on regions that align more closely with the semantics of the text.

\textbf{Quantitative analysis of the CI module.} As depicted in the final two columns of Fig.~\ref{fig:vis_feat}, the image features extracted by the visual encoder are highly abstract, devoid of semantically matching textual features. However, after integrating the CI module, the features have been significantly enhanced to align with the semantic attributes of the given text prompt. This signifies that the model is paying more attention to the areas described by the text. Additionally, for the fused feature \( V' \) in Fig.~\ref{fig:vis_feat}(e), we observed that its focus areas are consistent with the regions indicated by the high-weight words in \( W_{woi} \). This demonstrates that our visual and textual features are well-aligned.

\textbf{Quantitative analysis of \( W_{poi} \).} For \( W_{poi} \) in Fig.~\ref{fig:vis_feat}(b), the model adaptively focuses on the boundary regions between the foreground and background in the images. Notably, the lung boundary regions have higher weights. We believe this is because, in segmentation tasks, the pixels at the foreground-background boundaries are hard pixel cases, meaning they are the most challenging for the model to identify correctly. Our adaptive mask mechanism captures these hard pixel cases, masking and reconstructing them to better recognize these regions, demonstrating the effectiveness of our CVR module.

\textbf{Quantitative analysis of \( W_{woi} \).} For \( W_{woi} \) in Fig.~\ref{fig:vis_feat}(c), the model pays more attention to critical positional words like 'upper', 'left', 'middle', and 'lower'. Non-positional keywords, such as 'pulmonary', 'area', and 'lung', have relatively lower weights. This also proves the efficacy of our CLR module.

\textbf{Quantitative analysis of text-to-image understanding.} In Fig.~\ref{fig:chage_text}, we remove the text prompt indicating left lung infection and used a new prompt. The results in the last column show that the model doesn't predict the infection area in the left lung, but only in the right lung. This demonstrates that the model is well-aligned with the text, indicating it can understand the meaning of the text in the context of the image.

\section{Conclusion}
\label{sec: conclusion}
In this paper, we observe that the current methods for language-guided medical image segmentation yield results that are inconsistent with the semantics conveyed by the language, sometimes diverging significantly. Then we have proposed a Cross-Modal Conditioned Reconstruction for Language-guided Medical Image Segmentation (RecLMIS). This approach explicitly incorporates language and vision information as conditioning factors to facilitate contrastive learning and cross-modal reconstruction, which in turn contributes to achieving fine-grained visual-language alignment. Experiments on challenging medical semantic segmentation benchmarks demonstrate the outstanding accuracy of RecLMIS. More encouragingly, RecLMIS achieves state-of-the-art technology while significantly reducing parameter count and computational load. We will release the source code to the research community to promote the development of this emerging field.

\section{Limitations}
Although our work represents the most advanced research in the Language-guided Medical Image Segmentation (LMIS) field, there are still some limitations. We identify two main shortcomings in this work: 1. We find that the performance on the MosMedData+ test set is lower than on the QaTa-COV19 test set primarily because the former contains more small target samples. The model performs poorly on small targets, indicating room for improvement in this area. 2. Although word order has minimal impact on the model, we observed that synonyms do affect the output. We believe that employing large language models (LLMs) to increase data diversity and developing an image-sentence level contrastive loss can alleviate these issues in future work.

\bibliographystyle{ieeetr}

\end{document}